\documentclass[preprint,12pt]{elsarticle}

\usepackage{amssymb}
\usepackage{amsthm}

\newtheorem{assumption}{Assumption}

\usepackage{multirow}
\usepackage{caption}
\usepackage{amsmath}
\usepackage{booktabs}
\usepackage{subfigure}
\usepackage{color}

\journal{Neurocomputing}

\begin{document}
\begin{sloppypar}
\begin{frontmatter}



\title{Temporal Latent Variable Structural Causal Model for Causal Discovery under External Interferences}


\author[1,2]{Ruichu Cai}{}
\author[1]{Xiaokai Huang}{}
\author[1]{Wei Chen\corref{cor1}}
\ead{chenweidelight@gmail.com}
\author[1,3]{Zijian Li}{}
\author[1,4]{Zhifeng Hao}{}

\cortext[cor1]{Corresponding author}

\affiliation[1]{organization={School of Computer Science, Guangdong University of Technology},
            city={Guangzhou},
            country={China}}
\affiliation[2]{organization={Peng Cheng Laboratory},
            city={Shenzhen},
            country={China}}
\affiliation[3]{organization={Machine Learning Department, Mohamed bin Zayed University of Artificial Intelligence},
            city={Abu Dhabi},
            country={United Arab Emirates}}
\affiliation[4]{organization={College of Mathematics and Computer Science, Shantou University},
            city={Shantou},
            country={China}}

\begin{abstract}
Inferring causal relationships from observed data is an important task, yet it becomes challenging when the data is subject to various external interferences. Most of these interferences are the additional effects of external factors on observed variables. Since these external factors are often unknown, we introduce latent variables to represent these unobserved factors that affect the observed data. Specifically, to capture the causal strength and adjacency information, we propose a new temporal latent variable structural causal model, incorporating causal strength and adjacency coefficients that represent the causal relationships between variables. Considering that expert knowledge can provide information about unknown interferences in certain scenarios, we develop a method that facilitates the incorporation of prior knowledge into parameter learning based on Variational Inference, to guide the model estimation. Experimental results demonstrate the stability and accuracy of our proposed method.

\end{abstract}



\end{frontmatter}


\section{Introduction}
\label{Introduction}

Causal discovery~\cite{pearl2000models,pearl2009causality} from observed time series data is a significant task in various empirical sciences~\cite{ogburn2022causal,lagemann2023deep,li2023probabilities,li2023transferable}. Using causal discovery techniques can reveal the generation mechanism behind observational data, which helps decision-making, root cause analysis, etc. The typical model for causal discovery is the structural causal model (SCM), which implies the causal relationships between observed variables by functions~\cite{pearl2009causality}. Besides the SCM, the graphical model is another kind of model to represent the causal relationships. Many methods try to recover the causal graph with the acyclic assumption~\cite{chen2021causal,chen2022learning,cai2023learning,wang2023causal}. Based on these, time series data provides temporal information and also brings the challenge of the influence of time lag. Thus, the aim of causal discovery from time series data is to estimate the causal model and to recover the causal graph with temporal information.

Two typical approaches are proposed to address the aforementioned task: constraint-based methods and functional-based methods. Constraint-based methods, like PCMCI~\cite{runge2019detecting} and PCMCI+~\cite{runge2020discovering}, infer causal relationships between variables through (conditional) independence tests. While temporal data can offer time information to help determine causal direction, some equivalent classes may remain unidentified. On the other hand, functional-based methods model the data generation process using a Structural Causal Model, allowing for estimating the causal structure through model estimation techniques under some assumptions. For example, VAR-LiNGAM~\cite{hyvarinen2010estimation} relies on the non-Gaussian assumption and utilizes the Vector Autoregressive model to formalize the data generation process. Then, it can be estimated by applying the Independence Component Analysis (ICA). These methods focus on the problem of causal discovery with temporal information, without considering the variables in the data may be influenced by external interference.

\begin{figure}[t]
\centering 
    \includegraphics[width=0.9\columnwidth]{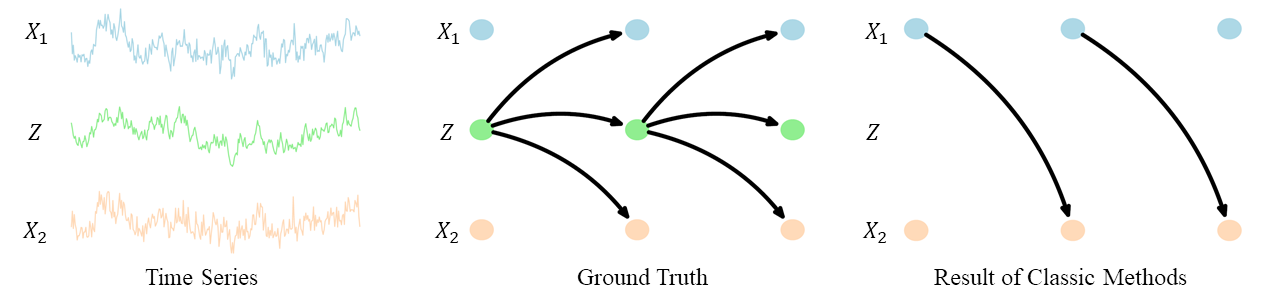}
\caption{$X_1$ and $X_2$ denote the fMRI data from brain region 1 and brain region 2. $Z$ denotes the unknown external interference, which can be regarded as a latent variable. Due to the existence of a latent variable, the result of classic methods contains spurious correlations.}
\label{fig:figure1}
\end{figure}

However, in real applications, the collected data may suffer from some unknown interferences. These ``interferences" denote the effects of external factors or environmental influences on observed variables. These external factors or environmental influences are always unmeasured or hidden, and their influence mechanisms remain unknown, complicating their incorporation into causal models. To address this challenge, we use latent variables to represent these unknown influences.
Inspired by this, we reformulate the task to learn causal structure from temporally observed data in the presence of latent variables. This task is tried to solve by utilizing different techniques. Based on the conditional independence tests method, LPCMCI algorithm~\cite{gerhardus2020high} is proposed as applicable to scenarios involving latent variables. Based on the data generation process, Geiger et al.~\cite{geiger2015causal} provide the identification theories of vector autoregressive processes with hidden components for causal discovery. It considers the causal relationship between two time series variables with a causal coefficient, which forms a causal matrix for all variables in the matrix form. Essentially, the presence of unknown external interference, which can be considered as environmental information, influences causal coefficients but also affects the adjacency of causal relationships, presenting challenges for causal discovery. Take Figure \ref{fig:figure1} as an example. For an individual, their brain regions may be affected by unknown external or environmental factors. These unknown factors influence the brain regions, which may lead to changes in the causal relationships between the brain regions. Additionally, external interference in a typical region could remove the causal parent on the target region, thereby altering the causal adjacency between the two regions. 
Building upon the previous analysis, we propose a causal model with latent variables by introducing causal weight and adjacency matrices to represent the relationships between variables using the vector autoregressive (VAR) model~\cite{canova1999vector}. In certain applications, external interference can be discerned through expert knowledge. Consequently, we estimate the model using variational inference~\cite{blei2017variational}, integrating expert knowledge or provided information on external interference as a prior. This approach enhances the precision of our model by incorporating relevant expertise and external influences during estimation.

Our contributions can be summarized as follows:

\begin{itemize}
    \item [1)]
    We address the challenge of causal discovery in the presence of external interference by the proposed Temporal Latent Variable Structural Causal Model. 
    \item [2)]
    We consider the causal matrix as a Hadamard product of an adjacency matrix and a weight matrix, which allows us to directly constrain the model’s complexity by adding a penalty term for the causal adjacency matrix during the optimization process and identify the causal relationships among observed variables more correctly.
    \item [3)] We estimate the Temporal Latent Variable Structural Causal model by employing variational inference, integrating expert knowledge or supplied information on external interference as a prior if applicable.
\end{itemize}

\section{Related Work}
\label{Related Work}
In this paper, we address the problem of causal discovery from time series data by accounting for latent variables that represent unknown external influences. So in this section, we investigate the related work of causal discovery from time series data with and without latent variables.

Traditional approaches for causal discovery always assume that the observed data follows the causal sufficient assumptions. That is, there is no hidden confounder~\cite{spirtes2000causation}. There are three main types of causal discovery methods for multiple time series. One type is the constraint-based method, such as PCMCI~\cite{runge2019detecting} and PCMCI+~\cite{runge2020discovering}. They perform a series of conditional independent tests to gain the Markov Equivalence Class of the causal structure. Unfortunately, such methods are time-consuming and challenging to scale to high-dimensional data due to the large number of tests required~\cite{spirtes2000causation}. Specifically, the computational complexity of constraint-based algorithms is similar to that of the PC algorithm~\cite{kalisch2007estimating}. In the worst case, the computational complexity of the PC algorithm is bounded by $O(p^q)$, where $p$ represents the number of nodes in the causal graph and $q$ denotes the maximal number of neighbors of all nodes. Furthermore, the experimental results in~\cite{bellotto2023enhancing} demonstrated that the time consumption of the PCMCI algorithm increases rapidly as the number of variables increases.
The second type is score-based methods, which search for the equivalence class with the highest score under some scoring criteria. The classical methods are GES\cite{chickering2002optimal}. For the time series data, DYNOTEARS~\cite{pamfil2020dynotears} is utilized for the estimation of both instantaneous and time-lagged relationships among variables. Its objective is to minimize penalized loss, ensuring that the resulting causal graph conforms to a Directed Acyclic Graph (DAG).
The third type is based on the structural causal model, which relies on the assumptions of specific causal functions and data distribution. The typical method, VARLiNGAM~\cite{hyvarinen2010estimation} extends the fundamental LiNGAM model to accommodate time series scenarios by integrating classic vector autoregressive models (VAR), which proves the identification of the linear causal model with non-Gaussian noises. To estimate the causal model, it provides a likelihood of the model and two practical methods, allowing for the analysis of both delayed and contemporaneous (instantaneous) causal relationships.


However, in practical application, due to data collection methods and costs, we can not always ensure that there is no confounder. To solve this problem, some researchers have tried to relax the causal sufficient assumption, and proposed several methods based on constraint or functional causal model~\cite{gerhardus2020high,entner2010causal,huang2015identification,chu2008search}. Based on the conditional independence tests, SVAR-FCI~\cite{malinsky2018causal} is proposed to identify the instantaneous and time-lagged effect in the presence of latent variables. The idea of SVAR-FCI is similar to that of PC and FCI~\cite{spirtes2000causation}, but it employs time invariance to deduce additional edge removals. Extending by PCMCI, Latent PCMCI (LPCMCI) algorithm allows for the existence of latent variables. LPCMCI uses variables' parents as default conditions to analyze effect sizes, avoiding inflated false positives and reducing the number of tested sets. Additionally, it introduces middle marks and LPCMCI-PAGs for clear causal interpretation and early edge orientation. In terms of the causal functional model, CLH-NV is a method that utilizes a first-order vector autoregressive process to handle latent components ~\cite{geiger2015causal}. It employs the EM algorithm to estimate the time-lagged linear effect between variables. Additionally, some methods try to fit the observed data with machine learning approaches to discover the Granger causality ~\cite{guo2008partial,yin2022deep,nauta2019causal}. But Granger causality is often used to determine predictive relationships between time series variables, but cannot definitively establish causal direction~\cite{shojaie2022granger}. It identifies statistical causation rather than actual causal mechanisms~\cite{granger1969investigating}. Therefore, we do not consider the Granger causality based method as our baseline. 

In summary, existing constraint-based methods primarily rely on conditional independence tests, which can be computationally intensive, especially when dealing with high-dimensional data and large sample sizes. Additionally, we note that function-based methods like CLH-NV, which use the EM algorithm, tend to suffer from overfitting. Granger causality-based methods, on the other hand, are more focused on time series prediction rather than uncovering causal mechanisms. In contrast, our proposed Temporal Latent Variable Structural Causal Model innovatively addresses the challenge of causal discovery in the presence of external interferences. By representing the causal matrix as a Hadamard product of an adjacency matrix and a weight matrix, our method can identify causal relationships among observed variables more correctly, differentiating our method from previous methodologies.

\section{Model Definition and Assumptions}
\label{Model Definition}

In this section, we provide a temporal latent variable structural causal model and assumptions for causal discovery under unknown interference. Formally, we are given the observed time series $\mathbf{X}=\{\mathbf X_1,\mathbf X_2,\cdots,\mathbf X_m\}$ that were generated from unknown interference. Let $\mathbf{Z}=\{\mathbf Z_1,\mathbf Z_2,\cdots,\mathbf Z_n\}$ denote the values of latent time series (or factors). 
Then at time $t$, the value of the observed variable $X_i$is denoted as $X_i(t)$ $i=1,2,\dots,m$, and the value of the latent factors $Z_i$ is denoted as $Z_i(t)$ for $i=1,2,\dots,n$, where $t=1,\cdots,T$. Following the structural vector autoregressive processes with a time lag $1$, the data generation processes of $X_i(t)$ and $Z_i(t)$ are formalized as:

\begin{equation}
    \begin{aligned}
        X_i(t) &=\sum_{j=1}^{m} C_{i,j}^{XX} X_j(t-1)+ \sum_{j=1}^{n}  C_{i,j}^{XZ} Z_j(t-1)+ N^{X_i}(t), \\
        Z_i(t)& =\sum_{j=1}^{n}  C_{i,j}^{ZX} X_j(t-1)+\sum_{j=1}^{m} C_{i,j}^{ZZ} Z_j(t-1)+ N^{Z_i}(t),
    \end{aligned}
    \label{eq: model_Xt}
\end{equation}
where $C_{i,j}^{XX}$ denote the causal matrix between observed variables $X_i$ and $X_j$. $C_{i,j}^{XZ}$ denote the causal effect of latent variable $Z_i$ on $X_j$. $N^{X_i}(t)$ and $N^{Z_i}(t)$ are two mutually independent noise terms for $X_{i}(t)$ and $Z_{i}(t)$, respectively. If for all noises of variables in $[\mathbf{X}, \mathbf{Z}]$ are independent of each other, the above structural vector autoregressive process is defined as \textit{diagonal structural VAR process}~\cite{geiger2015causal}. To clarify, we just consider the causal effect with a time lag $1$ in the above model, but note that our method can be extended to the case with a time lag that is more than $1$.

Considering that the latent variables imply external interference and influence the observed variable in the system, we assume that the latent variables are set as exogenous variables. Existing methods focus on estimating the causal matrix $\mathbf{C^{XX}}$ and $\mathbf{C^{XZ}}$ in Eq. (\ref{eq: model_Xt}). However, in practice, the external interference may affect the causal adjacent information or causal strength between two observed variables or both, we regard that the causal matrix can be composed of two parts: causal adjacency and causal strength. Thus, we incorporate the causal adjacency matrix and causal strength matrix to represent the causal relationship between two variables. Then the data-generation process of $X_i(t)$ and $Z_i(t)$ (where $t>1$) can be reformulated as:
\begin{equation}
\begin{aligned}
X_i(t) &=\sum_{j=1}^{m}A_{i,j}^{XX} \cdot W_{i,j}^{XX} X_j(t-1)+ \sum_{j=1}^{n} A_{i,j}^{XZ} \cdot  W_{i,j}^{XZ} Z_j(t-1)+ N^{X_i}(t), \\
Z_i(t)& =\sum_{j=1}^{m}A_{i,j}^{ZZ} \cdot W_{i,j}^{ZZ} Z_j(t-1)+ N^{Z_i}(t),
\end{aligned}
\end{equation}
where $A_{i,j}^{XX} \in \{0,1\}$ represents the existence of a causal edge from $X_j(t-1)$ to $X_i(t)$, and $W_{i,j}^{XX}$ represents the causal strength (or weight) from $X_j(t-1)$ to $X_i(t)$. The notations of $A_{i,j}^{XZ}$, $A_{i,j}^{ZZ}$, and $ W_{i,j}^{XZ}$, $W_{i,j}^{ZZ}$ are similar to $A_{i,j}^{XX}$ and $W_{i,j}^{XX}$, respectively. 

In terms of matrix, let $\mathbf{X}(t)\in \mathbb R^m$ denote the value of the observed time series at time $t$, and $\mathbf Z(t)\in \mathbb R^n$ denote the value of the unobserved ones, where $t=1,\cdots,T$. Then the above equations can be written as:
\begin{equation}
    \begin{aligned}
    \mathbf X(t) & =\mathbf{A}^{XX} \odot \mathbf{W}^{XX} \mathbf X(t-1)+\mathbf{A}^{XZ} \odot \mathbf{W}^{XZ} \mathbf Z(t-1)+\mathbf N^{X}(t), \\
    \mathbf Z(t) & =\mathbf{A}^{ZZ} \odot \mathbf{W}^{ZZ} \mathbf Z(t-1)+\mathbf N^{Z}(t),
    \end{aligned}
    \label{eq:model}
\end{equation}
where $\odot$ denotes Hadamard product, $\mathbf N^{X}(t)$ and $\mathbf N^{Z}(t)$ are noise terms which are independent with each other. 

From Equation (\ref{eq:model}), the causal matrix is the product of the causal adjacency matrix $\mathbf{A}$ and the causal weighted strength matrix $\mathbf{W}$. In detail, $\mathbf A=\begin{bmatrix}
\mathbf{A^{XX}}&\mathbf{A^{XZ}}\\\mathbf{0}&\mathbf{A^{ZZ}} 
\end{bmatrix}\in \{0,1\}^{(m+n)\times(m+n)}$ denotes the causal binary adjacency matrix, i.e., $A_{ij}=1$ means that the $j$-th time series influences the $i$-th one with time lag $1$. $\mathbf W=\begin{bmatrix}
    \mathbf{W^{XX}}&\mathbf{W^{XZ}}\\\mathbf{0}&\mathbf{W^{ZZ}} 
\end{bmatrix}\in \mathbb R^{(m+n)\times(m+n)}$ denotes the causal weighted strength matrix and $W_{ij}$ represents the causal effect from the $j$-th time series to the $i$-th one. Considering real-world applications, the extra interference is only influenced by itself at the previous timestamp. So we assumed that each $\mathbf Z_i,i=1,\cdots,n$, is an autoregressive time series that is only influenced by itself and independent of other latent variables. Therefore, $A^{ZZ}$ is a diagonal matrix and $A^{(ii)}W^{(ii)}\neq 0$ where $i=m+1,\cdots,m+n$. 

To ensure the identification of the proposed model and following the theoretical result in previous work~\cite{geiger2015causal}, it requires the following assumptions:
\begin{assumption}\label{asm:1}
    All noise $N^X_i,\,\,i=1,\cdots,m$ and $N^Z_j,\,\,j=1,\cdots,n$ follow non-Gaussian distribution.
\end{assumption}

\begin{assumption}\label{asm:2}
    The data generative process of our model is a diagonal-structural VAR process~\cite{geiger2015causal}. 
\end{assumption}

\begin{assumption}\label{asm:3}
    $\mathbf{A}^{XZ} \odot \mathbf{W}^{XZ}$ and $\mathbf M_1$ have full rank, where 
    \begin{equation}\label{equ:2}
    \mathbf M_1:=\mathbb E\left[
    \begin{pmatrix}
        \mathbf{X}(t)\\\mathbf{Z}(t)
    \end{pmatrix}
    (\mathbf X(t)^\mathrm{T},\mathbf X(t-1)^\mathrm{T})\right].
\end{equation} 
\end{assumption}

The model (\ref{eq:model}) is identifiable under Assumptions \ref{asm:1} - \ref{asm:3}, which is proven in~\cite{geiger2015causal}. In detail, the causal matrix among observed variables, $\mathbf{A}^{XX} \odot \mathbf{W}^{XX}$, can be uniquely identified only from observed data; the causal matrix from the latent variables to the observed variables, $\mathbf{A}^{XZ} \odot \mathbf{W}^{XZ}$, can be identified up to permutation and scaling. Note that the causal matrix among latent variables, $\mathbf{A}^{ZZ} \odot \mathbf{W}^{ZZ}$ is assumed to be a diagonal matrix.

Based on the proposed model given in Equation (\ref{eq:model}), our main task is to estimate the posteriors over parameters of the causal adjacency matrix $\mathbf{A}$, causal strength matrix $\mathbf{W}$ and latent variables $\mathbf{Z}$, given the observed variables $p(\mathbf{X})$. 

\section{Model Estimation}

Based on the aforementioned model of data generation processes, we design a method for model estimation based on variational inference.

\subsection{Marginal Likelihood}
Suppose that the observed time series data $\mathbf{X}$ are generated by Eq. (\ref{eq:model}), we aim to maximalize the marginal log-likelihood as 
\begin{equation}
    \begin{aligned}  
    \mathcal{L}(\mathbf{X}) & = \log \int \!\!\!\int \!\!\!\int p(\mathbf{X},\mathbf{Z},\mathbf{A},\mathbf{W}) d \mathbf{Z}d \mathbf{A} d \mathbf{W}\\
                            & = \log \int \!\!\!\int \!\!\!\int p(\mathbf{X} \mid \mathbf{Z}, \mathbf{A}, \mathbf{W})\times p(\mathbf Z,\mathbf A,\mathbf W)d \mathbf{Z}d \mathbf{A} d \mathbf{W}\\
=&\log \int \!\!\!\int \!\!\!\int p(\mathbf{X} \mid \mathbf{Z}, \mathbf{A}, \mathbf{W})\times p(\mathbf Z\mid\mathbf A,\mathbf W) \times p(\mathbf{A})\times p(\mathbf{W}) d \mathbf{Z}d \mathbf{A} d \mathbf{W}.
    \end{aligned}
\end{equation}

Then, given the parameters $\mathbf{A},\mathbf{W}$ and latent variables $\mathbf{Z}$, the marginal likelihood of observed data, denoted as $p(\mathbf X\mid \mathbf A,\mathbf W,\mathbf Z)$, which is calculated as 

\begin{equation}
\begin{aligned}
p(\mathbf{X} \mid \mathbf{A}, \mathbf{W}, \mathbf{Z}) & =\left[\prod_{t=2}^{T} p[\mathbf X(t) \mid \mathbf X(t-1), \mathbf{A}, \mathbf{W}, \mathbf Z(t-1)]\right]p[\mathbf X(1)]. 
\end{aligned}
\end{equation}

Based on the data generation process model as Eq. (\ref{eq:model}), we have $p[\mathbf X(1)]=p[\mathbf N^X(1)=\mathbf X(1)]$. The conditional likelihood $p[\mathbf X(t) \mid \mathbf X(t-1), \mathbf{A}, \mathbf{W}, \mathbf Z(t-1)]$, where $t\geq 2$, can be formalized as

\begin{equation}
\begin{aligned}
&p[\mathbf X(t) \mid \mathbf X(t-1), \mathbf{A}, \mathbf{W}, \mathbf Z(t-1)]\\
=&p\left[\mathbf N^X(t)=\mathbf X(t)-\left(\mathbf{A}^{XX} \odot \mathbf{W}^{XX} \mathbf X(t-1)+\mathbf{A}^{XZ} \odot \mathbf{W}^{XZ} \mathbf Z(t-1)\right)\right].
\end{aligned}
\end{equation}

The following problem is how to obtain $p(\mathbf Z\mid\mathbf A,\mathbf W)$, $p(\mathbf{A})$, $p(\mathbf{W})$ and $p(\mathbf{N^X}(t))$. Regarding $\mathbf{Z}$, $\mathbf{A}$, $\mathbf{W}$ and $\mathbf{N^X}(t)$ as latent variables, we will provide the prior over them in the following subsection.

\subsection{Prior over Latent Variables}

We assume the causal binary adjacency matrix and the causal weighted strength matrix among variables are random variables. Suppose that the prior distribution over $A$ is Bernoulli distributions, then we have

\begin{equation}
\begin{aligned}
p(\mathbf{A})&=\prod_{i=1}^{(m+n)} \prod_{j=1}^{(m+n)} p\left(A_{ij}\right),
\end{aligned}
\end{equation}
where $p\left(A_{ij}\right)=\operatorname{Bernoulli}(\rho)$ with $0\le\rho\leq1$. 

Suppose that the prior distribution over $W$ is Gaussian distribution, then

\begin{equation}
p\left(\mathbf{W}\right)=\prod_{i=1}^{(m+n)} \prod_{j=1}^{(m+n)} p\left(W_{ij}\right),
\end{equation}
where $p\left(W_{ij}\right)=\mathcal{N}\left(\mu^{w}, (\sigma^{w})^{2}\right)$ with mean $\mu^{w}$ and variance $(\sigma^{w})^{2}$.

Based on the above assumptions, the conditional distribution of the causal latent variables $\mathbf{Z}$ can be formalized as

\begin{equation}
\begin{aligned}
p(\mathbf Z\mid\mathbf A,\mathbf W)&=\left[\prod_{t=2}^{T} p[\mathbf Z(t) \mid \mathbf Z(t-1), \mathbf{A}, \mathbf{W}]\right]p[\mathbf Z(1)]. 
\end{aligned}
\end{equation}
with

\begin{equation}
\begin{aligned}
p[\mathbf Z(t)]=p[\mathbf N^Z(t)=\mathbf Z(t)],\text{ where }t=1,
\end{aligned}
\end{equation}

\begin{equation}
\begin{aligned}
&p[\mathbf Z(t) \mid \mathbf Z(t-1), \mathbf{A}, \mathbf{W}]
\\
=&p\left[\mathbf N^Z(t)=\mathbf Z(t)-\left(\mathbf{A}^{(ZZ)} \odot \mathbf{W}^{(ZZ)} \mathbf Z(t-1)\right)\right],\text{ where }t\geq 2.
\end{aligned}
\end{equation}

To guarantee universal applicability, we assume that noises $\mathbf N^{X}(t)$ and $\mathbf N^{Z}(t)$ follow Gaussian mixture distribution with $C$ components~\cite{calcaterra2008approximating}. Denote  $\mathbf s^X\in [1\cdots C]^{T\times m}$ and $\mathbf s^Z\in [1\cdots C]^{T\times n}$ as the indicator of which Gaussian distribution the noises of generative data is sampled from. Thus, for $t=1,\cdots,T$, the distribution of noises can be represented as

\begin{equation}
\begin{aligned}
p(N_i^X(t))&=\sum^C_{c=1}p(s^X_i(t)=c)p(N_i^X(t)\mid s^X_i(t)=c)\\
&=\sum^C_{c=1}\pi^X_{i,c}\mathcal N(N_i^X(t)\mid \mu^X_{i,c},(\sigma^X_{i,c})^2),\,\,\,\,\,i=1,\cdots,m,
\end{aligned}
\end{equation}

\begin{equation}
\begin{aligned}
p(N_i^Z(t))&=\sum^C_{c=1}p(s^Z_i(t)=c)p(N_i^Z(t)\mid s^Z_i(t)=c)\\
&=\sum^C_{c=1}\pi^Z_{i,c}\mathcal N(N_i^Z(t)\mid \mu^Z_{i,c},(\sigma^Z_{i,c})^2),\,\,\,\,\,i=1,\cdots,n,
\end{aligned}
\end{equation}

where

\begin{equation}
   \sum^C_{c=1}\pi^X_{i,c}=1,\mathbf\pi^X\in[0,1]^{m\times C},\mathbf\mu^X\in\mathbb R^{m\times C},\mathbf\sigma^X\in\mathbb R^{m\times C}, 
\end{equation}
and 
\begin{equation}
    \sum^C_{c=1}\pi^Z_{i,c}=1,\mathbf\pi^Z\in[0,1]^{n\times C},\mathbf\mu^Z\in\mathbb R^{m\times C},\mathbf\sigma^Z\in\mathbb R^{m\times C},
\end{equation}
are the parameters of the Gaussian mixture distribution.

\subsection{Variational Inference over Parameters}

The posterior $p(\mathbf{A}, \mathbf{W}, \mathbf{Z}\mid\mathbf{X})$ is intractable, so we use variational inference to approximate $p(\mathbf{A}, \mathbf{W}, \mathbf{Z}\mid\mathbf{X})$ with $q(\mathbf{A}, \mathbf{W}, \mathbf{Z})$. We assume $q(\mathbf{A}, \mathbf{W}, \mathbf{Z})$ is the distribution within the mean-field variational family, i.e.,

\begin{equation}
q(\mathbf Z,\mathbf A,\mathbf W)=q(\mathbf Z)\times q(\mathbf A)\times q(\mathbf W).
\end{equation}

Then the objective of the solution is to maximize the Evidence Lower Bound $\mathcal L_{elbo}$ as

\begin{equation}
\begin{aligned}
\mathcal L_{elbo}&=\mathbb E_{q(\mathbf Z,\mathbf A,\mathbf W)}\left[\ln p(\mathbf X\mid  \mathbf{A}, \mathbf{W}, \mathbf{Z})\right]+\mathbb E_{q(\mathbf Z,\mathbf A,\mathbf W)}\left[\ln \frac{p(\mathbf Z,\mathbf A,\mathbf W)}{q(\mathbf Z,\mathbf A,\mathbf W)}\right]\\
&=\mathcal L_{ell}+\mathcal L_{kl}.
\end{aligned}
\end{equation}

Inspired by the work in ~\cite{kingma2013auto,blei2017variational}, with the sparse constraint and hyper-parameter $\lambda$, we can rewrite the objective function as follows and aim to minimize:

\begin{equation}
\begin{aligned}
\mathcal L_{all} = -\mathcal L_{ell} + \lambda\|\mathbf{A}\|_1.
\label{eq: obj}
\end{aligned}
\end{equation}

To optimize Eq. (\ref{eq: obj}), we use the re-parameterization trick~\cite{blei2017variational, kingma2013auto} along with our approximate posterior as follows,

\begin{equation}
q(\mathbf{A})=\prod_{i=1}^{(m+n)} \prod_{j=1}^{(m+n)} q\left(A_{ij}\right), 
\end{equation}

\begin{equation}
\begin{aligned}
q(\mathbf{W})&=\prod_{i=1}^{(m+n)} \prod_{j=1}^{(m+n)} q\left(W_{ij}\right),
\end{aligned}
\end{equation}

\begin{equation}
\begin{aligned}
q(\mathbf Z)&=\prod^T_{t=1}\prod^m_{i=1}q[Z_{i}(t)],
\end{aligned}
\end{equation}
where $q\left(A_{ij}\right), q\left(W_{i j}\right)$ is the posterior distribution of $\mathbf{A}$ and $\mathbf{W}$. $\hat\rho_{ij}$, $\hat\mu^{w}_{ij}$, $\hat\sigma^{w}_{ij}$ is training parameters of our model. We set the approximate posterior distribution over $W_{i j}$ as $q\left(W_{i j}\right)=\mathcal{N}\left(\hat\mu^{w}_{ij}, 
(\hat\sigma^{w}_{ij})^{2}\right)$, which can be re-parameterized easily using $\hat\sigma^{w} = \hat\mu^{w}_{ij}+\hat\sigma^{w}_{ij}\mathbf{G}_w$, where $\mathbf{G}_w\sim\mathcal{N}(0,1)$. And since $\mathbf{A}$ is a binary matrix, The re-parameterization trick cannot be applied to  $\mathbf{A}$. Therefore, to guarantee differentiability, we model $q(\mathbf{A})$ using a continuous relaxation of the concrete distribution~\cite{jang2016categorical, maddison2016concrete} with relaxation parameter $\lambda_0$. In detail we set $q\left(A_{ij}\right) = \operatorname{Concrete}(\hat\rho_{ij},\lambda_0)$, i.e., 

$$
\begin{aligned}
&\mathcal{U}_A\sim Uniform(0,1),\\
&p_{ij} = \frac{\log\hat\rho_{ij} + \log\mathcal{U}_A - \log{(1-\mathcal{U}_A)}}{\lambda_0},\\
&A_{ij} = \frac{1}{1+\exp{(-p_{ij})}}.
\end{aligned}
$$

Then the derivatives of our objective with respect to the training parameters $\hat\Theta=(\hat\rho_{ij}, \hat\mu^{w}_{ij}, \hat\sigma^{w}_{ij})$ can be written as, 
$$
\nabla_{\hat\Theta}\mathcal L_{all}(\mathbf{X};\hat\Theta)=\nabla_{\hat\Theta}\mathcal L_{all}(g_{\hat\Theta}(\mathcal{U}_A, \mathbf{G}_w))=\mathcal L'_{all}(g_{\hat\Theta}(\mathcal{U}_A, \mathbf{G}_w))\nabla_{\hat\Theta}g_{\hat\Theta}(\mathcal{U}_A, \mathbf{G}_w),
$$where $g_{\hat\Theta}(\mathcal{U}_A, \mathbf{G}_w)$ describes the data generative process from $(\mathcal{U}_A, \mathbf{G}_w)$ to $\mathbf{X}$. 

\section{Experiments}
\label{Experiments}

\subsection{Synthetic Data}

In this section, to verify the performance of the proposed method, we carry out experiments on synthetic data generated according to Model (\ref{eq:model}).  

\textbf{Experiment setup.} To evaluate the sensitivity of the algorithm to different noise distributions, causal structures and sample sizes, We randomly generate noises $\mathbf N^X$ and $\mathbf N^Z$ from Gaussian mixture distribution with the number of components $C = 5$, uniform distribution from 0 to 1 and chi square distribution with degree of freedom of 2, respectively. The simulated datasets are generated according to different parameter settings: the number of samples $T = [500, \mathbf{1000}, 2500, 5000]$, the number of observed variables $m = [5, 10, \mathbf{20}, 25, 50, 100]$, the average in-degree of each node $d = [0.75, 1.0, \mathbf{1.25}, 1.5, 1.75]$ and the ratio of latent variables to observed variables $r = [0.2, 0.3, \mathbf{0.4}, 0.5, 0.6]$. For the causal adjacent and the causal strength, we independently generate $A_{ij}$ and $W_{ij}$ according to $p\left(A_{ij}\right)= \operatorname{Bernoulli}(\rho)$ where $\rho$ is the average in-degree, and
$p\left(W_{ij}\right) = \mathcal{N}\left(\mu^{w}, (\sigma^{w})^{2}\right)$ where $\mu^{w}$ is sampled from the uniform distribution $\mathcal{U}(0.5, 0.9)$ and $\sigma^{w}$ is sampled from $\mathcal{U}(0.001, 0.01)$, respectively. We randomly generate 10 datasets for each setting and conduct experiments on different datasets. And then, we calculate the average of the results of each experiment to obtain the final result of the algorithm. 

\textbf{Baselines.} To evaluate the effectiveness of our method, we use the following methods as baselines:

\begin{itemize}
    \item PCMCI~\cite{runge2019detecting}: It is a constraint-based method for evaluating the instantaneous and time-lagged effect between variables, by making use of the Momentary Conditional Independence (MCI) test. 
    \item VAR-LiNGAM~\cite{hyvarinen2010estimation}: This method is a function-based method to estimate a structural vector autoregression model. Based on the assumptions of causal sufficiency and non-Gaussian, it can identify the instantaneous and time-lagged linear effect between variables.
    \item SVAR-FCI~\cite{malinsky2018causal}: It is a constraint-based method based on FCI algorithm~\cite{spirtes2000causation} to identify the instantaneous and time-lagged effect in the presence of latent variables.
    \item CLH-NV~\cite{geiger2015causal}: It is a method based on one-order vector autoregression process designed to address latent components. It estimates the time-lagged linear effect between variables through the EM algorithm.
    \item LPCMCI~\cite{gerhardus2020high}: It is a constraint-based method, which extends PCMCI tailored to address latent variables, which improves the detection of causal relationships in autocorrelated time series by incorporating causal parents in conditioning sets.
    \item DYNOTEARS~\cite{pamfil2020dynotears}: It is a score-based method designed to estimate both instantaneous and time-lagged relationships between variables. The method is proposed based on NOTEARS~\cite{zheng2018dags} and minimizes the penalized loss to ensure that the estimated causal graph is a Directed Acyclic Graph (DAG). 
\end{itemize}

Note that much of the research on linear time series data primarily focuses on techniques that exclude latent variables. In these experiments, PCMCI and VAR-LiNGAM are employed as baseline methods to assess how methods that incorporate latent variables perform relative to those that do not. SVAR-FCI, LPCMCI and CLH-NV are three algorithms designed for causal discovery in the presence of latent variables. DYNOTEARS is a typical score-based method for learning the dynamic causal relationships from temporal data. The external interference may cause the causal relationships to change. We would like to use DYNOTEARS to verify whether external interference can be characterized by considering dynamic changes.

\textbf{Experiment metrics.} To compare the identified causal structures with different methods, we use \emph{Precision}, \emph{Recall} and \emph{F1 score} as the evaluation metrics. Based on ground truth, precision refers to the proportion of correctly learned edges to the total number of edges in the learned graph. Recall refers to the proportion of correctly learned edges to the total number of edges in the ground truth. F1 score is the balance index of precision and recall, which is calculated as  $2\times precision \times recall /(precision+recall)$. Additionally, we provide the standard deviations (SDs) of the Precision, Recall and F1 scores to show the variability of our results across different experiments and datasets.

\begin{figure}[t]
    \centering
    \begin{subfigure}
        \centering
        \includegraphics[scale=0.35]{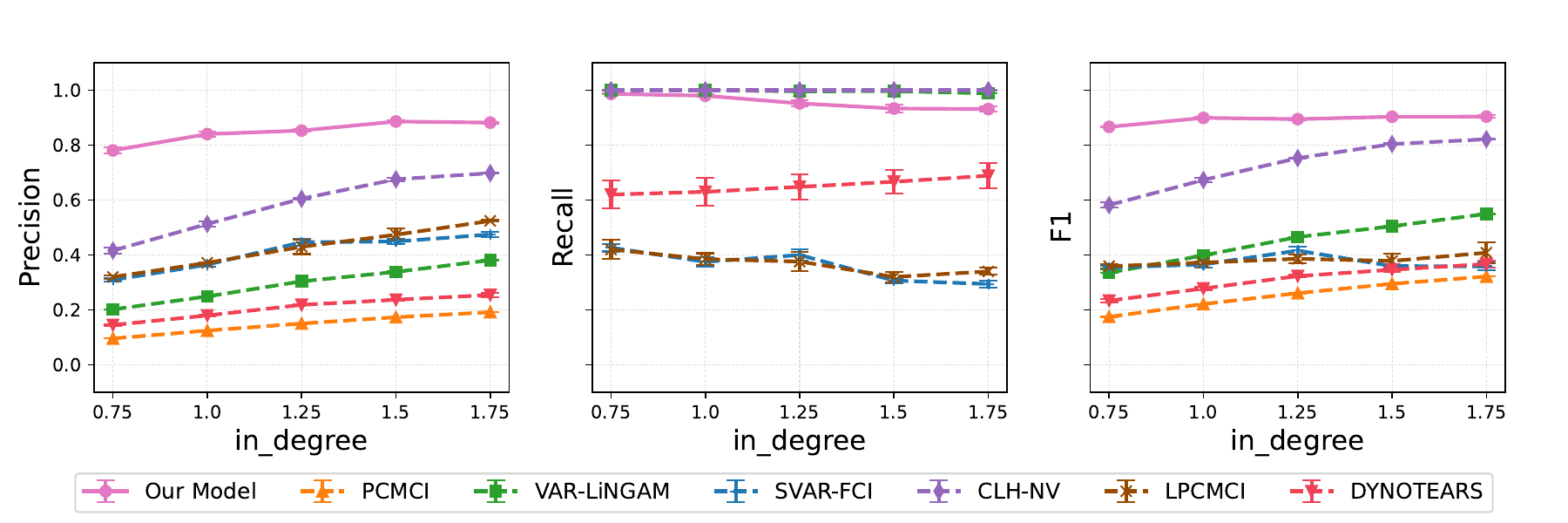}
        \par{\scriptsize (a) Precision, Recall and F1 score (with SDs) of Gaussian Mixture Distribution}
        \label{fig:in_degree1}
    \end{subfigure}
    \begin{subfigure}
        \centering
        \includegraphics[scale=0.35]{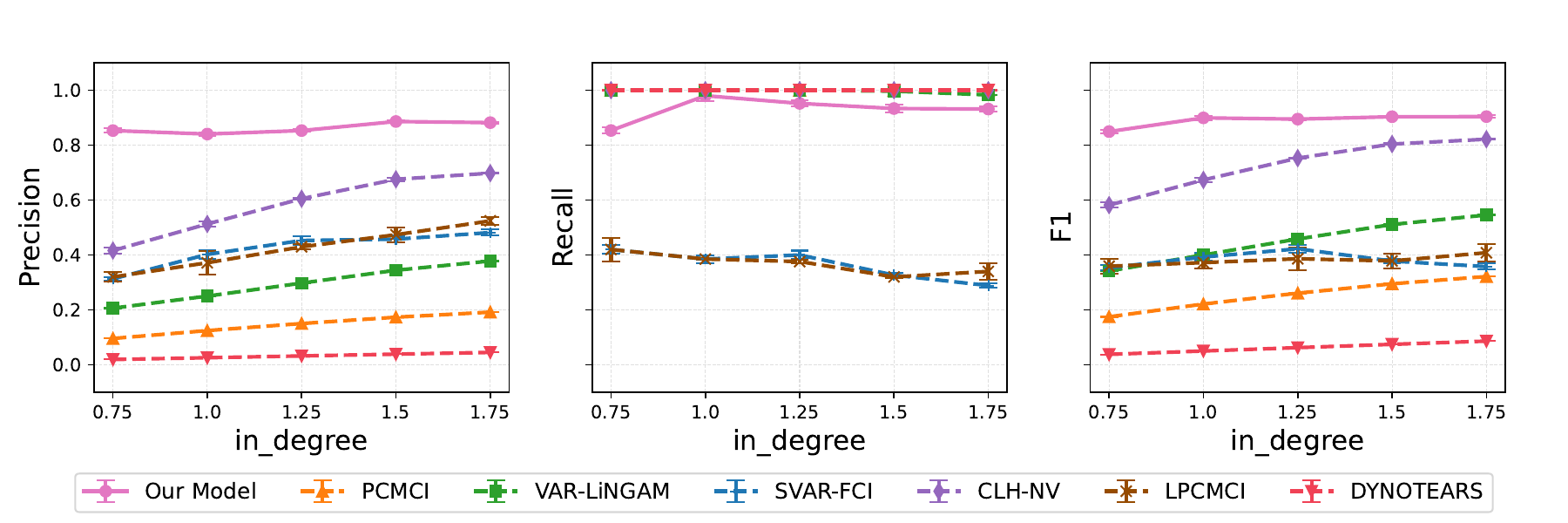}
        \par{\scriptsize (b) Precision, Recall and F1 score (with SDs) of Chi-square Distribution}
        \label{fig:in_degree2}
    \end{subfigure}
    \begin{subfigure}
        \centering
        \includegraphics[scale=0.35]{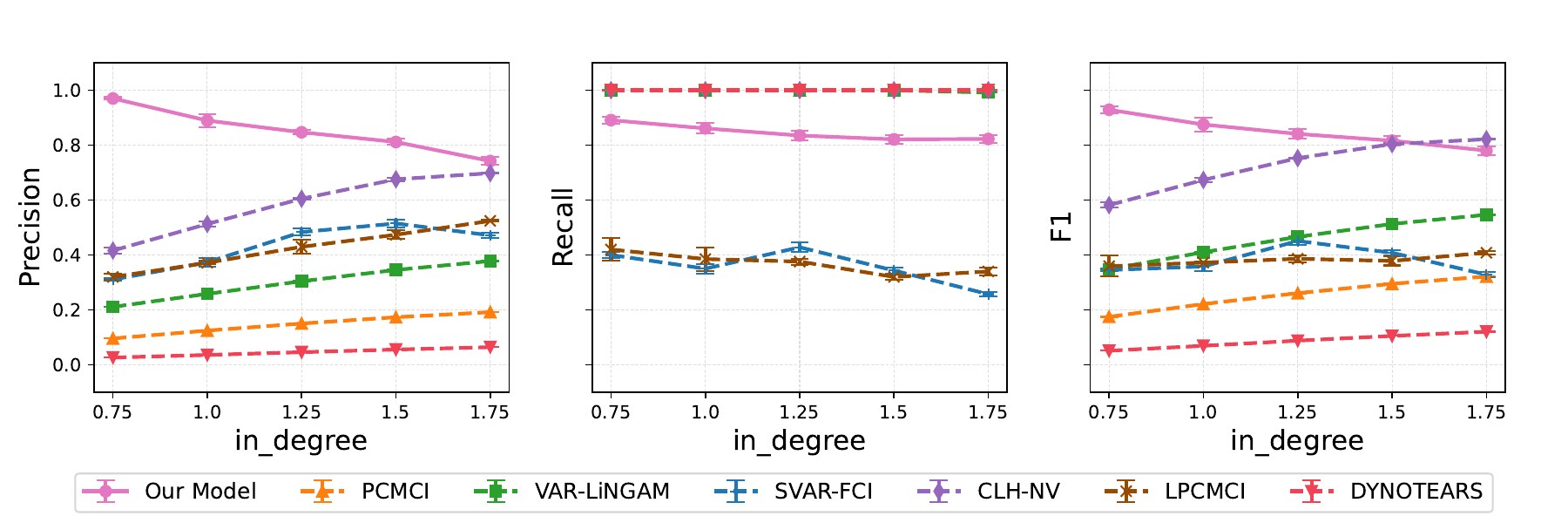}
        \par{\scriptsize (c) Precision, Recall and F1 score (with SDs) of Uniform Distribution}
        \label{fig:in_degree3}
    \end{subfigure}
    \caption{Results of different distributions with different average in-degrees of each node.}
    \label{fig:in_degree}
\end{figure}
\textbf{Sensitivity to the average in-degree of each node.} Figure \ref{fig:in_degree} illustrates the experimental results for different average in-degrees. In most cases, our method demonstrates superior precision values and F1 scores. The recall values for VAR-LiNGAM, CLH-NV, and our method are all close to 1. Despite their ability to identify most existing edges, their lower precision values suggest the presence of many redundant edges in their results due to unknown external interferences. Our method divides the causal matrix into an adjacency matrix and a weight matrix, reducing the impact of weights. These improvements aid in more accurately identifying observed variables affected by interferences, eliminating spurious correlations. Our method directly enforces sparsity constraints on the causal adjacency matrix and applies a uniform penalty across all edges. This approach mitigates the influence of causal edge weights on the sparsity constraints, thereby more effectively removing redundant edges. Although our recall is lower in certain cases, our method tends to remove redundant causal edges, and both our precision and F1 score are higher than those of other methods.

\begin{figure}[t]
    \centering
    \begin{subfigure}
        \centering
        \includegraphics[scale=0.35]{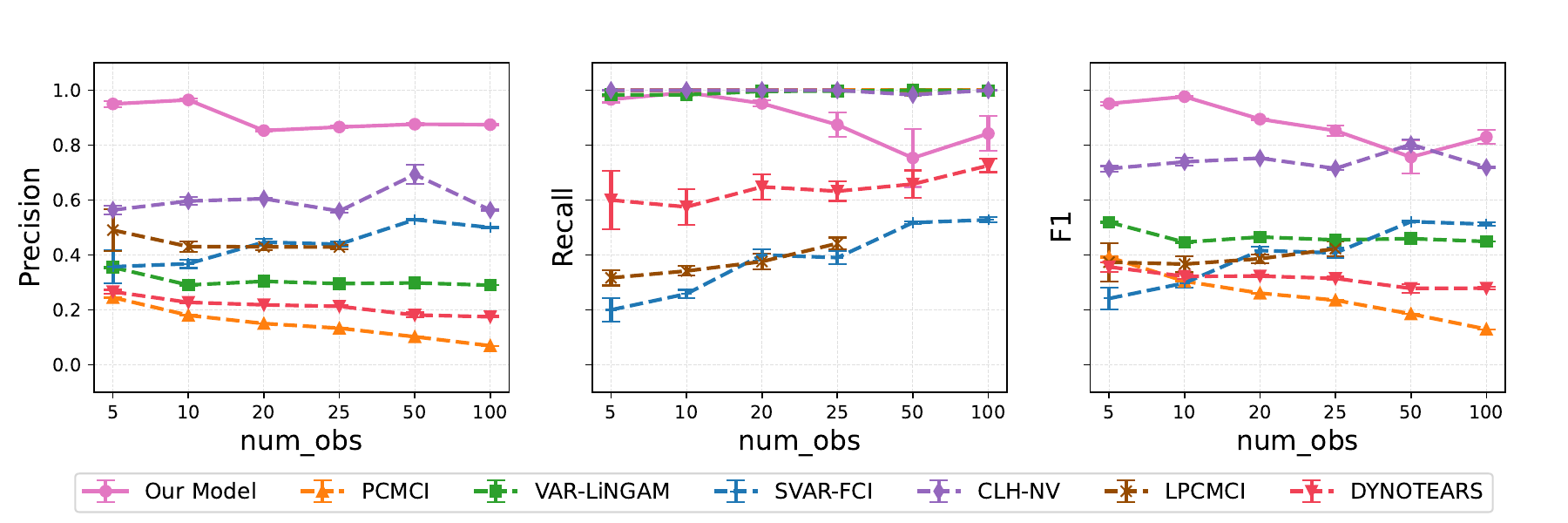}
         \par{\scriptsize (a) Precision, Recall and F1 score (with SDs) of Gaussian Mixture Distribution}
        \label{fig:num_obs1}
    \end{subfigure}
    \begin{subfigure}
        \centering
        \includegraphics[scale=0.35]{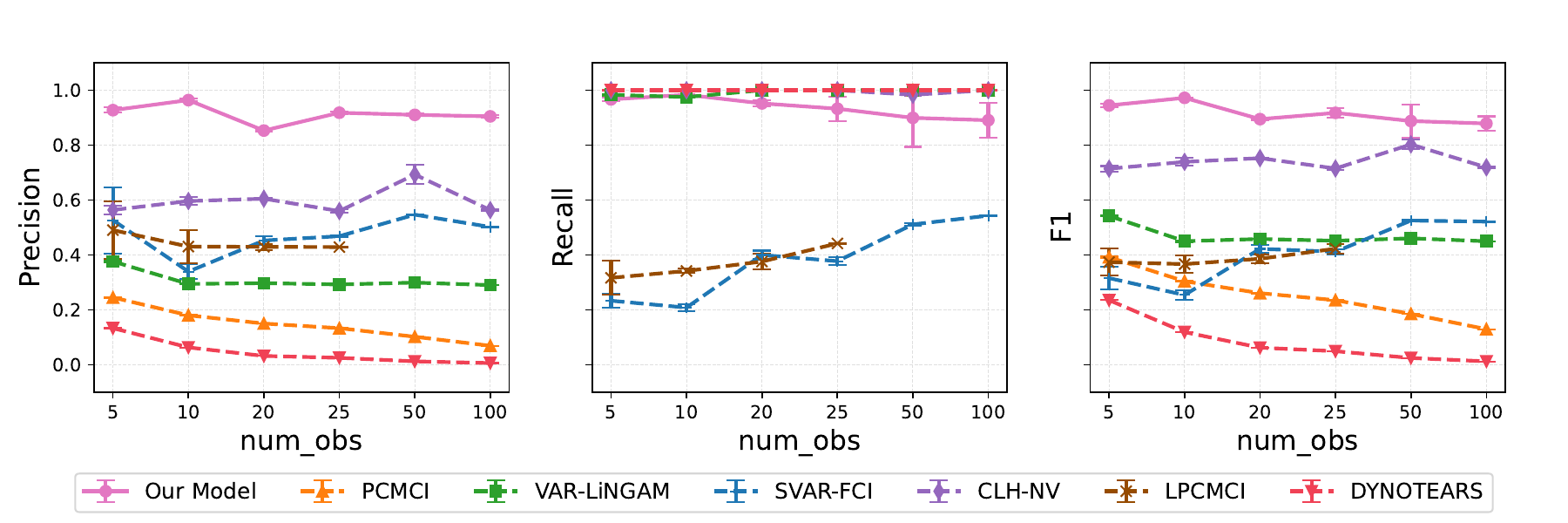}
        \par{\scriptsize (b) Precision, Recall and F1 score (with SDs) of Chi-square Distribution}
        \label{fig:num_obs2}
    \end{subfigure}
    \begin{subfigure}
        \centering
        \includegraphics[scale=0.35]{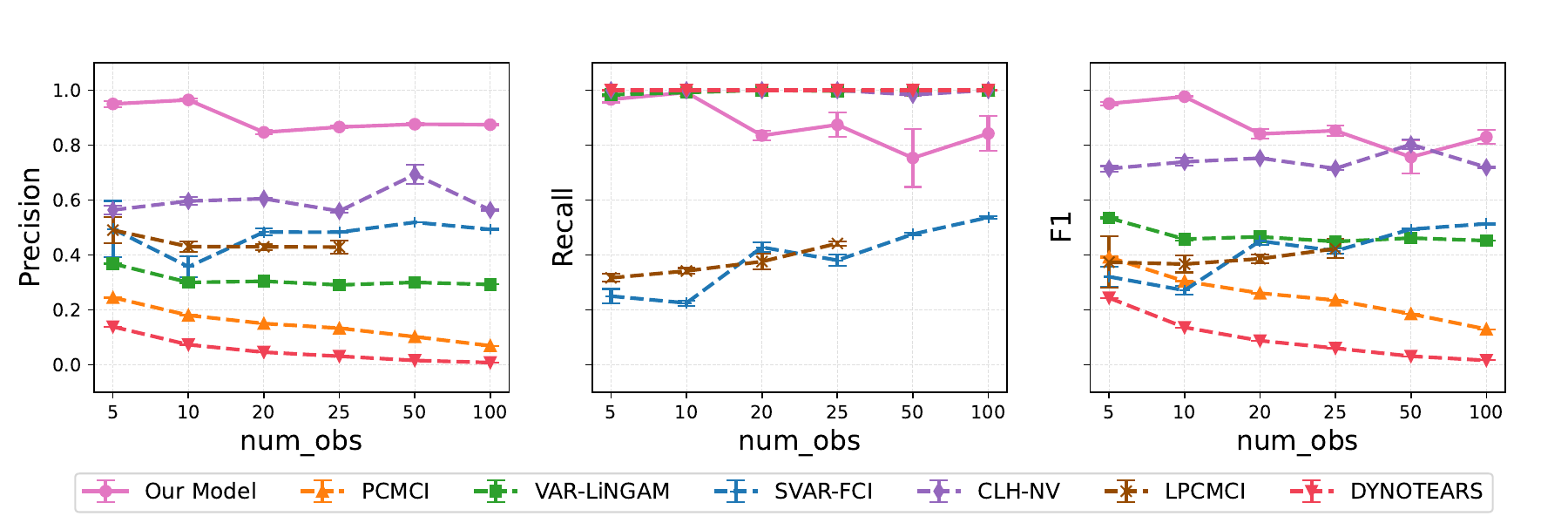}
        \par{\scriptsize (c) Precision, Recall and F1 score (with SDs) of Uniform Distribution}
        \label{fig:num_obs3}
    \end{subfigure}
    \caption{Results of different distributions with different numbers of observed variables.}
    \label{fig:num_obs}
\end{figure}

\textbf{Sensitivity to the number of observed variables.} Figure \ref{fig:num_obs} presents the experimental results across different numbers of observed variables. The exponential time complexity of LPCMCI makes it impractical for high-dimensional data ($> 25$ variables), with runs exceeding one month. Thus, we only report results completed within one month. From the results, it is observed that the precision and F1 score of other methods are notably inferior to our method. However, When the number of observed variables exceeds 20, the recall for PCMCI, VAR-LiNGAM, and CLH-NV are all close to 1, while our method experiences a decrease in recall. Additionally, DYNOTEARS, SVAR-FCI and LPCMCI exhibit an increasing trend in recall. Conversely, the precision of baselines decreases significantly with the growing number of nodes. This indicates that, although an increase in the number of nodes enables them to identify more correct edges, they also become more susceptible to spurious correlations. This issue arises from the interaction between poorly calibrated conditional independence (CI) tests for a smaller number of variables, which inflate false positives due to autocorrelation, and the sequential testing approach for a larger number of variables, which reduces false positives.

\begin{figure}[t]
    \centering
    \begin{subfigure}
        \centering
        \includegraphics[scale=0.35]{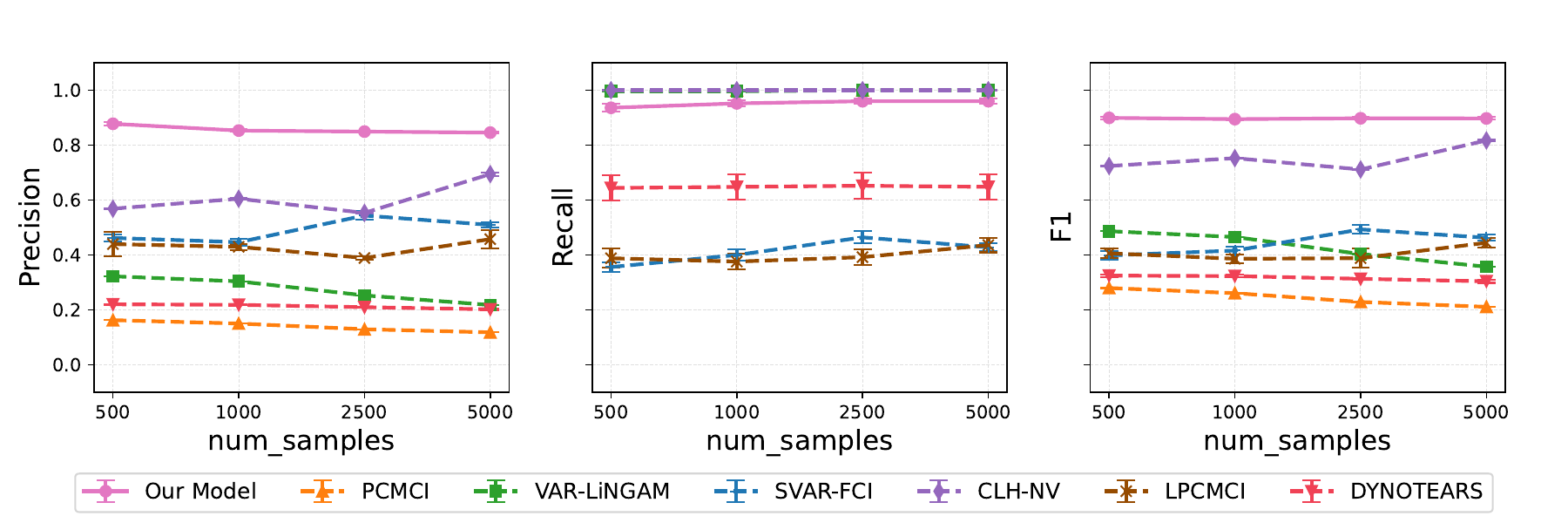}
        \par{\scriptsize (a) Precision, Recall and F1 score (with SDs)
         of Gaussian Mixture Distribution}
        \label{fig:num_samples1}
    \end{subfigure}
    \begin{subfigure}
        \centering
        \includegraphics[scale=0.35]{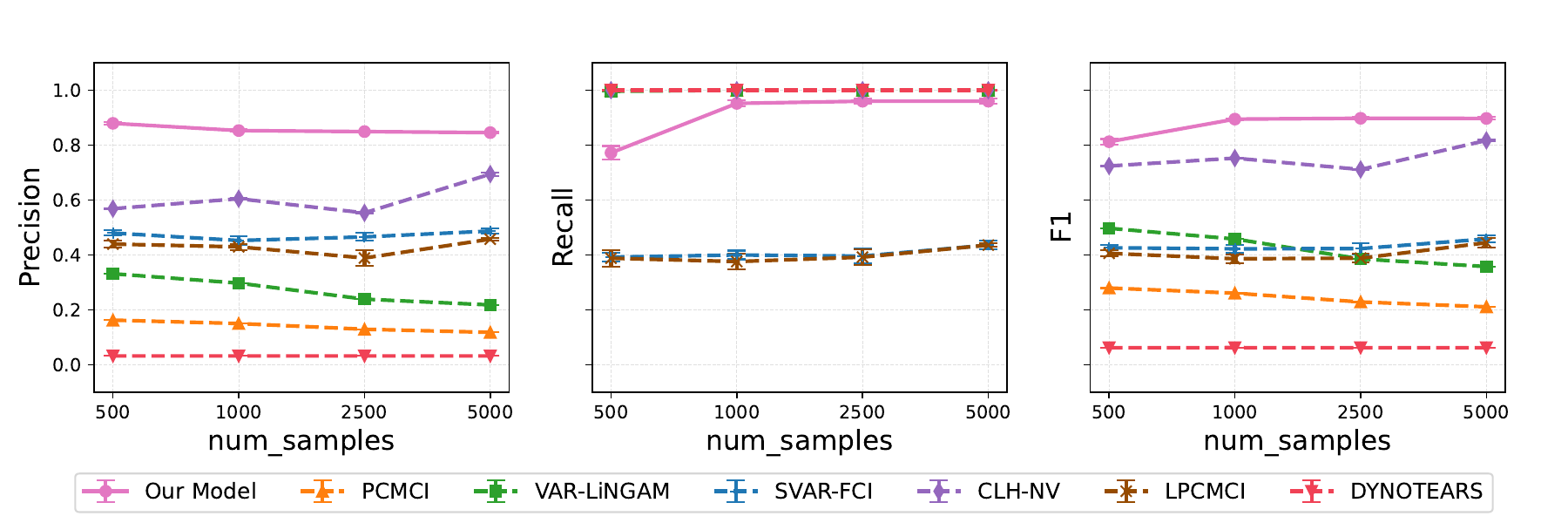}
        \par{\scriptsize (b) Precision, Recall and F1 score (with SDs) of Chi-square Distribution}
        \label{fig:num_samples2}
    \end{subfigure}
    \begin{subfigure}
        \centering
        \includegraphics[scale=0.35]{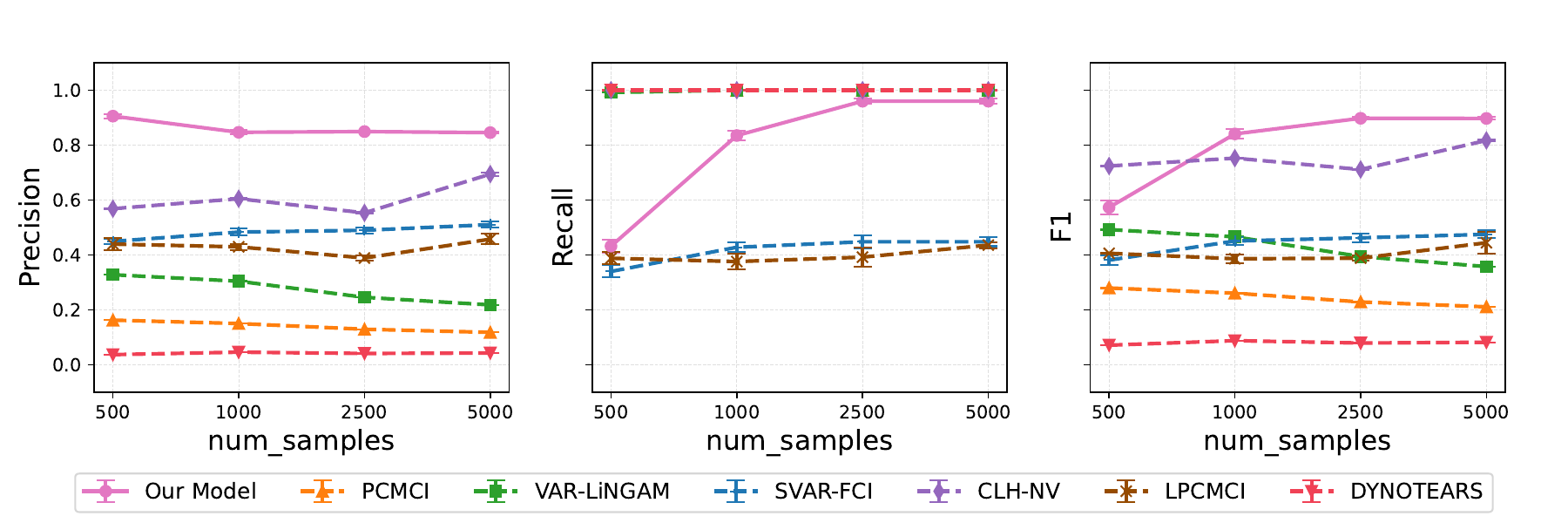}
        \par{\scriptsize (c) Precision, Recall and F1 score (with SDs) of Uniform Distribution}
        \label{fig:num_samples3}
    \end{subfigure}
    \caption{Results of different distributions with different sample sizes.}
    \label{fig:num_samples}
\end{figure}

\textbf{Sensitivity to the sample size.} Figure \ref{fig:num_samples} illustrates the experimental results across varying sample sizes. Notably, our method exhibits significantly higher precision compared to other methods. This can be attributed to the separation of the adjacency matrix and weight matrix, enabling us to mitigate the influence of causal weight and the intensity of the influence of external factors. As a result, we can effectively identify targets for external interference and mitigate the impact of spurious correlations on estimation. Furthermore, it is observed that the precision, recall and F1 score of our method consistently maintain stability and remain at a relatively high level across different sample sizes. Increasing the sample size provides more data for conditional independence tests, which leads to a slight improvement in the performance of SVAR-FCI and LPCMI. The variations in sample size have minimal impact on the performance of DYNOTEARS. Additionally, we observe that when the sample size is larger or equal than $1000$, the algorithms' performances do not improve much even with large increases in the sample size. This can be attributed to two reasons. First, the presence of hidden variables imposes an upper limit on the performance of baseline methods. Second, data quality significantly impacts model performance. high-quality samples can significantly enhance accuracy, while noisy samples hinder it. According to statistical learning theory, a high level of noise in samples restricts the model's learning ability and limits improvements in accuracy, even with larger sample sizes~\cite{kap2021effect}. 

\begin{figure}[t]
    \centering
    \begin{subfigure}
        \centering
        \includegraphics[scale=0.35]{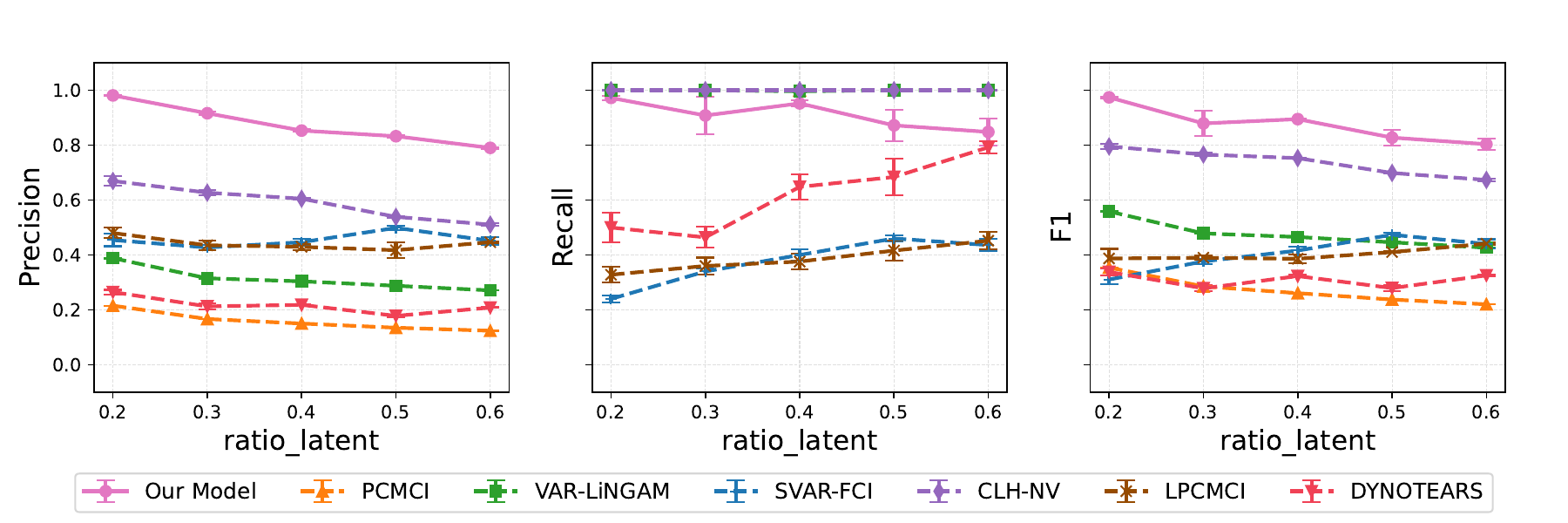}
        \par{\scriptsize (a) Precision, Recall and F1 score (with SDs) of Gaussian Mixture Distribution}
        \label{fig:ratio_latent1}
    \end{subfigure}
    \begin{subfigure}
        \centering
        \includegraphics[scale=0.35]{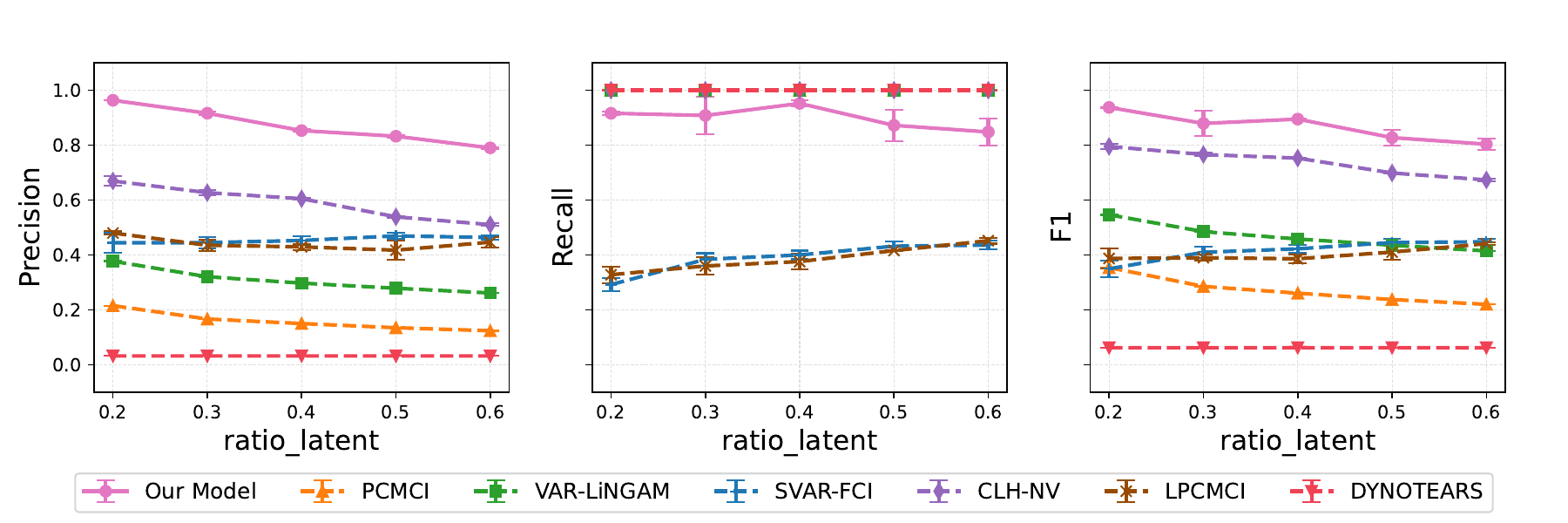}
        \par{\scriptsize (b) Precision, Recall and F1 score (with SDs) of Chi-square Distribution}
        \label{fig:ratio_latent2}
    \end{subfigure}
    \begin{subfigure}
        \centering
        \includegraphics[scale=0.35]{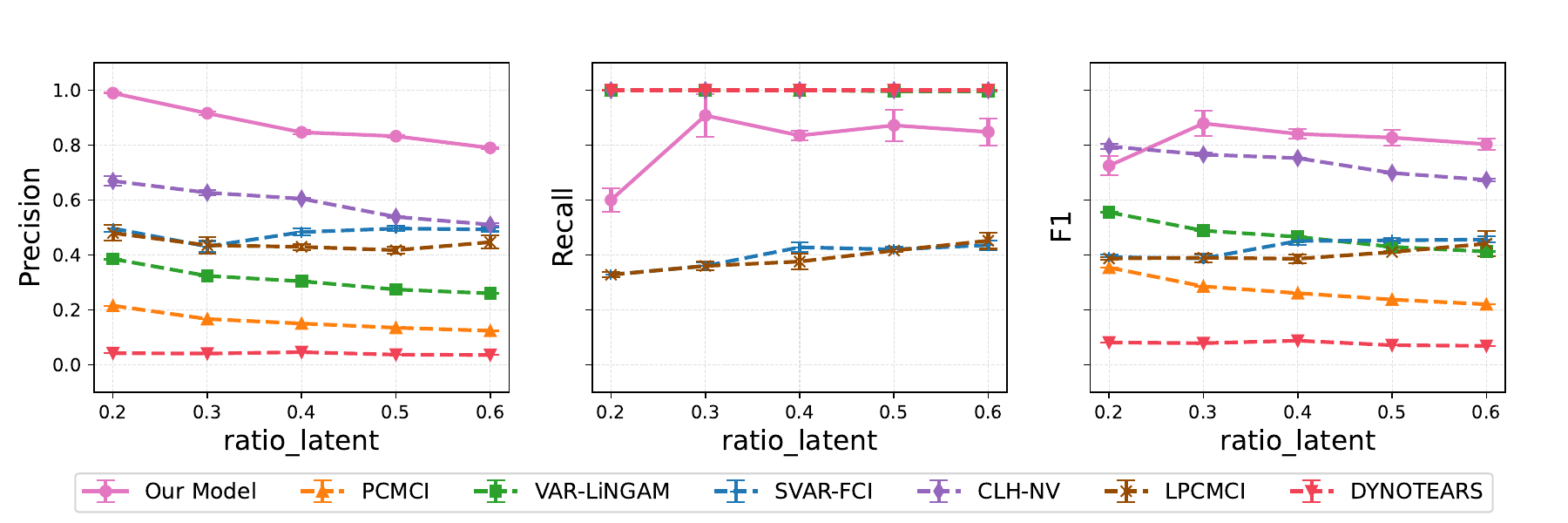}
        \par{\scriptsize (c) Precision, Recall and F1 score (with SDs) of Uniform Distribution}
        \label{fig:ratio_latent3}
    \end{subfigure}
    \caption{Results of different distributions with different ratios of latent variables.}
    \label{fig:ratio_latent}
\end{figure}

\textbf{Sensitivity to the ratio of latent variables to observed variables.} Figure \ref{fig:ratio_latent} presents the experimental results across different latent variable ratios. It can be observed that as the ratio of latent variables increases, the recall of other methods either increases or remains at a high level. However, their precision and F1 scores are much lower than those of our method. As the number of variables increases, cumulative errors can arise during the learning of causal structures or optimization of the objective function, which affects the performance of baseline methods. The recall and precision of our method consistently remain at a relatively high level. This indicates that the introduction of prior information on latent variables can help reduce the impact of spurious correlations, thereby eliminating redundant edges.


\subsection{fMRI data}

To test the performance of our method in practical application, we perform our method on functional magnetic resonance imaging (fMRI) dataset~\cite{smith2011network}. This dataset contains realistic, simulated BOLD data for 28 different underlying causal networks. Considering the assumption of our model, we applied our method to \textit{sim2} and \textit{sim6}. Each dataset consists of 50 individual data. \textit{Sim2} has 10 nodes, and its fMRI session for each subject is 10 min. \textit{Sim6} has 10 nodes, and its fMRI session for each subject is 60 min. To align with our setup, we selected different sets of exogenous variables in the ground truth as unobserved variables, with set sizes of 1, 2, and 3, while the remaining variables were treated as observable variables.

The results are given in Table \ref{tab:1} and Table \ref{tab:2}. Our method obtains the highest precision and F1 score. Although the recall of our method is not the highest, the causal edges learned by our method are largely correct. The recall of DYNOTEARS is the highest, while its precision is very small. This indicated that its results may be overfitting. Compared to CLH-NV, our method incorporates a sparsity penalty on the causal adjacent matrix, which uniformly affects the probability of the existence of each causal edge. This approach helps reduce the likelihood of overfitting.

\hspace*{-10cm}
\begin{table}[t]\footnotesize
    \flushleft
    \scalebox{0.65}{
    \begin{tabular}{ll|ccccccc}
        \hline
          $N$ & {Metrics} & Our method & PCMCI & VAR-LiNGAM & SVAR-FCI & CLH-NV & LPCMCI & DYNOTEARS \\
        \hline
          {} & Precision & $\mathbf{0.947\pm 0.093}$ & ${0.224\pm 0.028}$ & ${0.39\pm 0.053}$ & ${0.403\pm 0.136}$ & ${0.395\pm 0.098}$ & ${0.405\pm 0.173}$ & ${0.232\pm 0.017}$ \\
          {1} & Recall & ${0.467\pm 0.049}$ & ${0.569\pm 0.053}$ & ${0.516\pm 0.07}$ & ${0.151\pm 0.048}$ & ${0.723\pm 0.305}$ & ${0.16\pm 0.069}$ & $\mathbf{0.882\pm 0.059}$ \\
          {} & F1 score & $\mathbf{0.621\pm 0.042}$ & ${0.321\pm 0.034}$ & ${0.442\pm 0.05}$ & ${0.216\pm 0.063}$ & ${0.511\pm 0.273}$ & ${0.226\pm 0.092}$ & ${0.367\pm 0.025}$ \\
         \hline
          {} & Precision & $\mathbf{0.978\pm 0.052}$ & ${0.241\pm 0.029}$ & ${0.415\pm 0.06}$ & ${0.436\pm 0.189}$ & ${0.415\pm 0.051}$ & ${0.404\pm 0.182}$ & ${0.232\pm 0.017}$ \\
          {2} & Recall & ${0.479\pm 0.058}$ & ${0.588\pm 0.058}$ & ${0.539\pm 0.071}$ & ${0.148\pm 0.055}$ & ${0.698\pm 0.072}$ & ${0.161\pm 0.076}$ & $\mathbf{0.878\pm 0.071}$ \\
          {} & F1 score & $\mathbf{0.641\pm 0.057}$ & ${0.341\pm 0.035}$ & ${0.467\pm 0.056}$ & ${0.215\pm 0.072}$ & ${0.521\pm 0.064}$ & ${0.226\pm 0.099}$ & ${0.382\pm 0.028}$ \\
         \hline
          {} & Precision & $\mathbf{0.893\pm 0.084}$ & $0.283\pm 0.041$ & $0.480\pm 0.077$ & $0.414\pm 0.202$ & $0.407\pm 0.048$ & $0.394 \pm 0.11$ & $0.145\pm 0.003$ \\
          {5} & Recall & $0.566\pm 0.039$ & $0.612\pm 0.063$ & $0.561\pm 0.073$ & $0.145\pm 0.068$ & $0.716\pm 0.081$ & $0.140 \pm 0.05$ & $\mathbf{0.995\pm 0.018}$ \\
          {} & F1 score & $\mathbf{0.680\pm 0.050}$ & $0.386\pm 0.043$ & $0.514\pm 0.065$ & $0.208\pm 0.092$ & $0.516\pm 0.053$ & $0.217 \pm 0.054$ & $0.253\pm 0.005$ \\
         \hline
    \end{tabular}
    }
    \caption{The results on fMRI data (sim 2), where $N$ is the number of latent variables.}
    \label{tab:1}
\end{table}

\hspace*{-10cm}
\begin{table}[t]\footnotesize
    \flushleft
    \scalebox{0.65}{
    \begin{tabular}{ll|ccccccc}
        \hline
          $N$ & {Metrics} & Our method & PCMCI & VAR-LiNGAM & SVAR-FCI & CLH-NV & LPCMCI & DYNOTEARS \\
        \hline
          {} & Precision & $\mathbf{0.984\pm 0.053}$ & ${0.195\pm 0.023}$ & ${0.383\pm 0.045}$ & ${0.402\pm 0.098}$ & ${0.583\pm 0.11}$ & ${0.387\pm 0.11}$ & ${0.299\pm 0.033}$ \\
          {1} & Recall & ${0.466\pm 0.038}$ & ${0.75\pm 0.073}$ & ${0.614\pm 0.064}$ & ${0.225\pm 0.064}$ & ${0.701\pm 0.232}$ & ${0.234\pm 0.09}$ & $\mathbf{0.866\pm 0.068}$ \\
          {} & F1 score & $\mathbf{0.63\pm 0.031}$ & ${0.309\pm 0.033}$ & ${0.471\pm 0.048}$ & ${0.284\pm 0.071}$ & ${0.61\pm 0.178}$ & ${0.288\pm 0.094}$ & ${0.444\pm 0.04}$ \\
         \hline
          {} & Precision & $\mathbf{1.0\pm 0.0}$ & ${0.206\pm 0.022}$ & ${0.408\pm 0.042}$ & ${0.411\pm 0.101}$ & ${0.553\pm 0.215}$ & ${0.388\pm 0.124}$ & ${0.316\pm 0.035}$ \\
          {2} & Recall & ${0.479\pm 0.032}$ & ${0.767\pm 0.075}$ & ${0.652\pm 0.079}$ & ${0.224\pm 0.074}$ & ${0.725\pm 0.215}$ & ${0.22\pm 0.096}$ & $\mathbf{0.871\pm 0.071}$ \\
          {} & F1 score & $\mathbf{0.647\pm 0.03}$ & ${0.325\pm 0.032}$ & ${0.5\pm 0.047}$ & ${0.286\pm 0.081}$ & ${0.627\pm 0.114}$ & ${0.276\pm 0.104}$ & ${0.463\pm 0.043}$ \\
         \hline
          {} & Precision & $\mathbf{0.939\pm 0.052}$ & $0.266\pm 0.031$ & $0.474\pm 0.043$ & $0.435\pm 0.120$ & $0.603\pm 0.051$ & $0.44 \pm 0.143$ & $0.148\pm 0.003$ \\
          {5} & Recall & $0.568\pm 0.025$ & $0.800\pm 0.086$ & $0.672\pm 0.079$ & $0.235\pm 0.086$ & $0.731\pm 0.075$ & $0.262 \pm 0.104$ & $\mathbf{0.995\pm 0.018}$ \\
          {} & F1 score & $\mathbf{0.700\pm 0.033}$ & $0.399\pm 0.043$ & $0.554\pm 0.048$ & $0.298\pm 0.093$ & $0.653\pm 0.047$ & $0.321 \pm 0.112$ & $0.257\pm 0.005$ \\
         \hline
    \end{tabular}
    }
    \caption{The results on fMRI data (sim 6), where $N$ is the number of latent variables.}
    \label{tab:2}
\end{table}



\subsection{Financial Time Series}

We also apply our algorithm to financial market data~\cite{kleinberg2013causality}, which contains 25 stock portfolios. From this data, we choose 7 datasets with a time lag of 1. Each dataset consists of 4000-day time periods. At each temporal instance $t$, a portfolio is contingent upon the values of the three factors at that specific time plus an error term. We randomly select 1, and 2 exogenous variables as latent variables, with the remaining variables treated as observed variables. Due to LPMCI's inability to produce results within a month for financial data, we present only the comparative results between our proposed method and other baseline approaches, as detailed in Table \ref{tab:3}.

As shown in Table \ref{tab:3}, although CLH-NV exhibits high recall, their precision and F1 scores are much lower than ours. This implies that their result contains many redundant edges due to the presence of latent variables, while our method can identify these redundant edges. This can be interpreted that the sparse constraint on the adjacency matrix allows us to learn a more compact causal structure. Furthermore, VAR-LiNGAM, PCMCI and DYNOTEARS obtain low precision and F1 scores. A potential reason for this could be the presence of external irregular factors influencing the financial time series data, rendering these two methods ineffective. SVAR-FCI also obtains low precision, recall and F1 score, which implies that its results contain many undetermined causal edges. These edges indicate that there might be latent variables or existing Markov equivalent classes.

\hspace*{20cm}
\begin{table}[t]\footnotesize
    \centering
    \flushleft
    \scalebox{0.75}{
    \begin{tabular}{ll|cccccc}
        \hline
          $N$ & {Metrics} & Our method & PCMCI & VAR-LiNGAM & SVAR-FCI & CLH-NV & DYNOTEARS \\
        \hline
          {} & Precision & $\mathbf{0.815\pm 0.128}$ & ${0.011\pm 0.008}$ & ${0.018\pm 0.011}$ & ${0.011\pm 0.038}$ & ${0.331\pm 0.023}$ & ${0.031\pm 0.033}$ \\
          {1} & Recall & ${0.711\pm 0.119}$ & ${0.420\pm 0.127}$ & ${0.327\pm 0.153}$ & ${0.015\pm 0.006}$ & $\mathbf{0.951\pm 0.051}$ & ${0.098\pm 0.087}$ \\
          {} & F1 score & $\mathbf{0.752\pm 0.094}$ & ${0.021\pm 0.015}$ & ${0.033\pm 0.020}$ & $0.0127\pm 0.015$ & ${0.491\pm 0.031}$ & ${0.043\pm 0.040}$ \\
         \hline
          {} & Precision & $\mathbf{0.740\pm 0.044}$ & $0.025\pm 0.010$ & $0.086\pm 0.034$ & $0.010\pm 0.018$ & $0.356\pm 0.088$ & $0.008\pm 0.015$ \\
          {2} & Recall & $0.643\pm 0.066$ & $\mathbf{1.000\pm 0.000}$ & $0.988\pm 0.016$ & $0.016\pm 0.025$ & $0.966\pm 0.021$ & $0.214\pm 0.364$ \\
          {} & F1 score & $\mathbf{0.664\pm 0.041}$ & $0.049\pm 0.020$ & $0.156\pm 0.057$ & $0.011\pm 0.020$ & $0.506\pm 0.098$ & $0.016\pm 0.027$ \\
         \hline
    \end{tabular}
    }
    \caption{The results on financial data, where $N$ is the number of latent variables.}
    \label{tab:3}
\end{table}

\section{Conclusion}

In this study, we present a new temporal latent causal model for identifying the causal structure in the presence of varying external interference. In the proposed model, latent variables are employed to adeptly characterize unobserved external interference. The intricate nature of diverse interference is encapsulated through the formulation of causal adjacency and causal strength matrices, enabling the representation of both hard and soft interference.  Furthermore, to enhance the adaptability of the method and accommodate expert knowledge, we employ a variational inference-based approach for model estimation, leveraging such knowledge as priors. Our experimental results validate the effectiveness and robustness of the proposed model and algorithms. In further work, we will focus on generalizing the model to accommodate non-linear scenarios, to make it more applicable.

\section*{Acknowledgements}

This research was supported in part by National Science and Technology Major Project of China (2021ZD0111501), the National Science Fund for Excellent Young Scholars (62122022), National Natural Science Foundation of China (U24A20233, 62206064), the Guangzhou Basic and Applied Basic Research Foundation (2024A04J4384), and the Guangdong Basic and Applied Basic Research Foundation (2025A1515010172, 2023B1515120020).




 \bibliographystyle{elsarticle-num-names} 
 \bibliography{sample}




\end{sloppypar}
\end{document}